%% file: main.tex
\documentclass[conference]{IEEEtran}
\usepackage[pdftex]{graphicx}
\usepackage{amsmath,amssymb,amsfonts,multirow}
\usepackage{algorithmic}
\usepackage{tikz}
\usepackage{subfigure}
\usepackage{stfloats}
\usepackage{makecell}
\usepackage{siunitx}

\usepackage[acronym, nonumberlist]{glossaries}
\input{acronyms.tex}
\newcommand*\Bell{\ensuremath{\boldsymbol\ell}}

\glsdisablehyper

\makeatletter
\newcommand{\linebreakand}{%
  \end{@IEEEauthorhalign}
  \hfill\mbox{}\par
  \mbox{}\hfill\begin{@IEEEauthorhalign}
}
\makeatother

\usepackage[draft]{changes}
\definechangesauthor[name={Hemanth Manjunatha}, color=purple]{HM}

\begin{document}
\title{Beyond One Model Fits All: Ensemble Deep Learning for Autonomous Vehicles}

\author{
	\IEEEauthorblockN{Hemanth Manjunatha}
	\IEEEauthorblockA{School of Aerospace Engineering \\
		Georgia Institute of Technology \\
		Atlanta, Georgia \\
		\texttt{hmanjunatha6@gatech.edu}}
	\and
	\IEEEauthorblockN{Panagiotis Tsiotras}
	\IEEEauthorblockA{School of Aerospace Engineering\\
	Georgia Institute of Technology}
	Atlanta, Georgia \\
	\texttt{tsiotras@gatech.edu}}

\maketitle
\begin{abstract}
	Deep learning has revolutionized autonomous driving by enabling vehicles to perceive and interpret their surroundings with remarkable accuracy. This progress is attributed to various deep learning models, including Mediated Perception, Behavior Reflex, and Direct Perception, each offering unique advantages and challenges in enhancing autonomous driving capabilities. However, there is a gap in research addressing integrating these approaches and understanding their relevance in diverse driving scenarios. This study introduces three distinct neural network models corresponding to Mediated Perception, Behavior Reflex, and Direct Perception approaches. We explore their significance across varying driving conditions, shedding light on the strengths and limitations of each approach. Our architecture fuses information from the base, future latent vector prediction, and auxiliary task networks, using global routing commands to select appropriate action sub-networks. We aim to provide insights into effectively utilizing diverse modeling strategies in autonomous driving by conducting experiments and evaluations. The results show that the ensemble model performs better than the individual approaches, suggesting that each modality contributes uniquely toward the performance of the overall model. Moreover, by exploring the significance of each modality, this study offers a roadmap for future research in autonomous driving, emphasizing the importance of leveraging multiple models to achieve robust performance.
\end{abstract}

\input{sections/introduction}

\input{sections/related_works}

\input{sections/approach}

\input{sections/results}

\input{sections/conclusion}

\bibliographystyle{IEEEtran}
\bibliography{references}
\end{document}

%% file: acronyms.tex
\newacronym{lstm}{LSTM}{Long-Short Term Memory}
\newacronym{gru}{GRU}{Gated Recurrent Unit}
\newacronym{rnn}{RNN}{Recurrent Neural Network}
\newacronym{sac}{SAC}{Soft Actor-Critic}
\newacronym{vae}{VAE}{Variational Autoencoder}
\newacronym{bev}{BEV}{Bird-Eye View}
\newacronym{cnn}{CNN}{Convolutional Neural Network}
\newacronym{rl}{RL}{Reinforcement Learning}
\newacronym{ddpg}{DDPG}{Deep Deterministic Policy Gradient}
\newacronym[longplural={Artificial Neural Networks}]{ann}{ANN}{Artificial Neural Network}
\newacronym{mdn}{MDN}{Mixture Density Network}
\newacronym{gan}{GAN}{Generative Adversarial Network}
\newacronym{mlp}{MLP}{Multi-Layer Perceptron}
\newacronym{imu}{IMU}{Inertial Measurement Unit}
\newacronym{dqn}{DQN}{Deep Q-Network}
\newacronym{mse}{MSE}{Mean Squared Error}
\newacronym{msssim}{MS-SSIM}{Multi-Scale Structural Similarity Index Measure}
\newacronym{a2c}{A2C}{Advantage Actor Critic}
\newacronym{ppo}{PPO}{Proximal Policy Optimization}
\newacronym{ae}{AE}{Autoencoder}
\newacronym{dae}{DAE}{Dynamic Autoencoder}
\newacronym{cdae}{CDAE}{Combined Dynamic Autoencoder}
\newacronym{CARNet}{CARNet}{\textbf{C}ombined dyn\textbf{A}mic autoencode\textbf{R} \textbf{Net}work}
\newacronym{KARNet}{KARNet}{\textbf{K}alman Filter \textbf{A}ugmented \textbf{R}ecurrent Neural \textbf{Net}work}
\newacronym{dvae}{DVAE}{Dynamic Variational Autoencoder}
\newacronym{wm}{WM}{World Models}
\newacronym{mdnrnn}{MDN-RNN}{Mixture Density Recurrent Neural Network}
\newacronym{mpc}{MPC}{Model-Predictive Control}
\newacronym{il}{IL}{Imitation Learning}
\newacronym{dl}{DL}{Deep Learning}
\newacronym{mpn}{MPN}{Mediated Perception Network}
\newacronym{brn}{BRN}{Behavior Reflex Network}
\newacronym{dpn}{DPN}{Direct Perception Network}
\newacronym{apn}{APN}{Action Perception Network}

%% file: sections/introduction.tex
\section{Introduction}
Deep learning has played a pivotal role in propelling the development of autonomous driving. 
Over the past years, deep learning techniques involving training neural networks on massive datasets have enabled vehicles to perceive and interpret their surroundings with unprecedented accuracy~\cite{grigorescu2020survey}. 
Notably, deep learning models with structure (for example, world models~\cite{ha2018world}) have emerged as a crucial approach in advancing autonomous driving capabilities~\cite{hafner2019dream, ma2021contrastive}. 
By learning a good representation through structured deep learning models, vehicles can effectively process complex information from various sensors and sources, enhancing their perception and decision-making abilities. 
To date, these models can be broadly categorized into three overarching paradigms: Mediated Perception~\cite{ullman1980against}, Behavior Reflex~\cite{pomerleau2012neural}, and Direct Perception approaches~\cite{chen2015deepdriving}.
Nevertheless, there still remains a lack of research addressing the amalgamation of all three approaches into a single model and answering whether the individual approaches maintain their salience under varying driving conditions. 
This study aims to remedies these issue by introducing three distinct neural network models corresponding to the Mediated Perception, Behavior Reflex, and Direct Perception approaches, thereby elucidating the significance of each approach across diverse driving scenarios.

\textbf{Mediated Perception}: It involves constructing a \textit{World Model} by understanding different elements in the scene such as lanes, other cars, and traffic lights~\cite{janai2020computer, ullman1980against}. 
By \textit{World Model}, we mean architectures that are designed explicitly to acquire internal models of the environment~\cite{ha2018world}. 
Indeed, evidence from recent neuroscience/cognitive science research \cite{downing2009predictive, svensson2013dreaming} supports the idea of constructing internal models of the environment, i.e., World Model (WM), to predict the consequences of the actions is a natural way to achieve desired interaction of the agent with its surroundings.
The predictive WM, typically, involves an \glspl{ae} and a \glspl{rnn} to deduce low-dimensional ``latent variables" from data with temporal correlations \cite{ha2018recurrent, lipton2015critical, salehinejad2017recent, plebe2020road, manjunatha2023karnet}.
Even though constructing the World Model has benefited driving~\cite{werbos1987learning, silver2017predictron}, comprehending the entire scene might introduce unnecessary intricacy. 
Moreover, not all the objects in a scene are relevant for driving. 
Thus, the Mediated Perception can suffer from learning redundant representations while adding more complexity.

\begin{figure}[!ht]
	\centering
	\includegraphics[width=\linewidth]{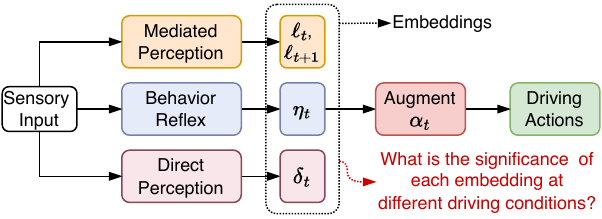}
	\caption{Overview of the ensemble deep learning architecture for autonomous driving.}
	\label{fig:overview}
\end{figure}

\textbf{Behavior Reflex}: Directly maps the input sensory input (usually images) to driving actions~\cite{huang2023overview, chen2015deepdriving, chib2023recent}. 
The mapping of sensory input to the driving action is usually achieved by imitation learning, where a network is trained (in a regression/classification fashion) using the data collected by an expert driver~\cite{le2022survey}. 
Even though the approach is straightforward, there are several drawbacks. First, for the same sensory input, different expert drivers can take drastically different actions, which leads to an ill-posed problem~\cite{zheng2022imitation}. 
For example, consider a scenario where a pedestrian suddenly steps onto the road. A driver might brake, swerve, or continue driving cautiously depending on factors like speed, distance, road conditions, and traffic density. 
A Behavior Reflex model would struggle to discern between these situations, and will likely produce inconsistent or incorrect responses. Second, the Behavior Reflex model's performance lies in the granularity of the driving actions they generate. 
These models tend to have low-level or high-level actions, making learning and adaptation difficult. 
Generally, Behavior Reflex models are often designed to produce specific motor responses, such as turning the steering wheel by a fixed angle or applying the brakes with a predetermined force.
However, driving is a continuous and dynamic process where the appropriate actions vary widely in intensity and duration.
For instance, smoothly decelerating when approaching a red light requires nuanced control, which a rigid low-level Behavior Reflex model might struggle to achieve. 
Nonetheless, the Behavior Reflex can provide a good baseline model, which can be improved by augmenting other models, such as Mediated Perception and Direct Perception.

\textbf{Direct Perception}: Falls in between Mediated Perception and Behavior Reflex models. 
Instead of learning World Models by parsing the entire sensory input or by directly mapping the sensory input to driving action, Direct Perception maps the sensory information to intermediate \textit{affordance} indicators of road conditions that influence driving decisions~\cite{chen2015deepdriving}. 
The affordance vector can then be used to learn appropriate driving actions.
Nonetheless, the direction perception might not be enough for all driving conditions, and it might suffer from the same drawbacks as Behavior Reflex (e.g., low-level driving action) and Mediated Perception. 
In such situations, an ensemble model using all three approaches can provide more information (which might be redundant) to make the driving action more robust.

\begin{figure*}[!ht]
	\centering
	\includegraphics[width=\textwidth]{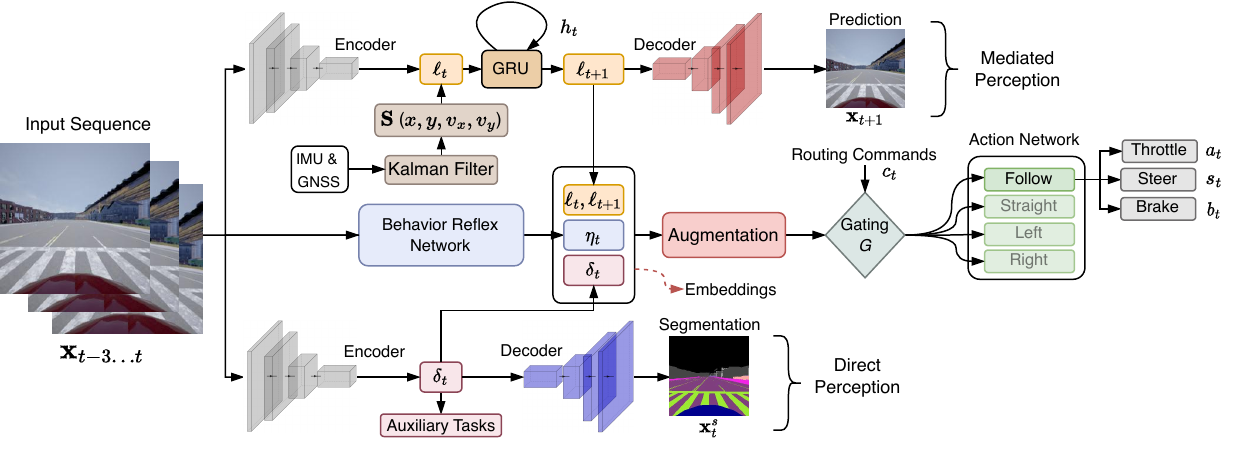}
	\caption{Overall architecture inspired by \cite{liang2018cirl}. The architecture consists of five sub-networks: Behavior Reflex network, Mediated Perception network, Direct Perception network, action prediction network, and value network. The Mediated Perception network and Direct Perception network are trained offline, and the parameters are frozen during \gls{il} training. The embeddings from the Behavior Reflex network, Mediated Perception network, and Direct Perception network are augmented to form the input for the action prediction network. The network uses the routing commands from a global path planner to select a sub-network in action prediction. Each sub-network in action prediction corresponds to following the lane, turn left, turn right, and moving forward. The output of these sub-networks are two waypoints, which are converted to throttle, steer, and brake using a PID controller.}
	\label{fig:overall-architecture}
\end{figure*}

To study the effect of an ensemble model, we introduce a neural network architecture that explicitly combines redundant information in a structured manner as shown in Fig.~\ref{fig:overall-architecture}. 
The network architecture is inspired by \cite{liang2018cirl}. The architecture consists of four sub-networks: a \textit{Behavior Reflex network, a Mediated Perception network, a Direct Perception network, and an Action Prediction network}. 
The implementation details of each network are discussed in Section \ref{sec:proposed_method}. 
The Mediated Perception network and Direct Perception network are trained offline. 
Only the Behavior Reflex and Action Prediction networks are trained using imitation learning. 
All three networks receive a stack of three images ($\mathbf{x}_{t-3}\dots \mathbf{x}_t$) at time instance $t$. 
The Behavior Reflex network compresses the input images to an embedding $\eta_t$ using a ResNet. 
The Mediated Perception network learns the embedding $\Bell_{t}$ and $\Bell_{t+1}$, which signifies the present and future state of the traffic. 
The Direct Perception network embeds the input images to $\delta_t$ through convolution operation while simultaneously predicting auxiliary tasks such as distance to the front vehicle and traffic light conditions.
The reasoning for using auxiliary tasks is to provide rich information (e.g., information about traffic lights, crossings, obstacles) and images to better learn the actions. 
These four embeddings ($\Bell_{t+1},\ \eta_t,\ \text{and}\ \delta_t$) are augmented together to form a feature vector that is used by the Action Prediction network. 
Further, to support goal-oriented actions, we introduce a gating unit that selects different branches of the action network depending on the global routing commands \cite{liang2018cirl}. 
It is crucial to emphasize that the primary focus of this research is not the development of novel neural network architecture but rather the investigation and comprehension of diverse modeling strategies in various driving scenarios. This study aims to provide valuable insights into the practicality of employing a range of models, effectively serving as a roadmap to guide research efforts in the context of end-to-end deep learning for autonomous driving.

%% file: sections/related_works.tex
\section{Related Works}
In autonomous driving applications using a Mediated Perception approach, the agent recognizes relevant objects~\cite{teichmann2018multinet, huval2015empirical} in the scene by processing the sensory data and constructs a world model/representation~\cite{okamoto2019vision, chen2015deepdriving}. 
This world model can then be used to derive appropriate driving actions. The sensory inputs for vision-based autonomous driving for instance involve images; however, other sensors such as IMU, GNSS, Odemeter, and Lidar sensors can be used~\cite{yeong2021sensor}. 
Appropriate neural network architectures with multiple processing pipelines should be used to fuse the information from different sensors~\cite {wang2020multi}.
Along this direction, physics infused neural network architectures are gaining a lot of traction where the neural networks leverage the structure of a physics model and data-driven learning capabilities of the deep neural networks~\cite{manjunatha2023karnet, wormann2022knowledge, zhou2020kfnet, barratt2020fitting, girin2021dynamical}. 
Despite the successful application of the Mediated Perception in autonomous driving, some challenges exist. 
For example, a Mediated Perception model must process sensory inputs that vary in complexity and processing methodology.
Nevertheless, in the context of coexisting methodologies, the influence-Mediated Perception approach warrants further scrutiny.

Instead of constructing an explicit world model like Mediated Perception, the Behavior Reflex model directly maps the sensory input (in vision-based autonomous driving, it is usually images) to driving actions~\cite{huang2023overview, chen2015deepdriving, chib2023recent}. One common technique used in Behavior Reflex is imitation learning~\cite{pan2018agile, le2022survey}, where the autonomous vehicle learns by imitating the behavior of human drivers. 
This involves training a neural network using data collected from human-driven vehicles. 
While imitation learning is straightforward, it has some limitations. First, different human drivers may respond differently to the same situation, making it challenging to determine the correct action solely based on human demonstration data. 

Direct Perception, conversely, occupies an intermediate position between Mediated Perception and Reflexive Behavior~\cite{chen2015deepdriving}. 
Instead of building a complete World Model or directly mapping the input to driving actions, Direct Perception seeks to extract essential ``affordance" indicators from sensor data~\cite{lee2021deep}. 
In this direction, Chen et al.~\cite{chen2015deepdriving} introduced 14 indicators (including heading angle, distances to preceding cars, lane markings in a three-lane highway, and a "fast" Boolean), using camera images with AlexNet+14 for autonomous driving.
Al-Qizwini et al.~\cite{al2017deep} built upon DeepDriving, proposing five indicators (heading angle and distances to lane markings) while removing the five distance indicators to preceding cars from~\cite{chen2015deepdriving}. 
Sauer et al.~\cite{sauer2018conditional} expanded the Direct Perception approach to include high-level driving commands, suggesting six affordance indicators (heading angle, distance to the vehicle ahead, distance to lane centerline, red light, speed sign, and hazard stop) for complex urban environments.

In this section, we have delved into various approaches to autonomous driving, each offering a unique perspective and set of solutions to the complex challenges of self-driving vehicles. 
Mediated Perception emphasizes constructing comprehensive world models from sensory data and has demonstrated considerable success in autonomous driving.
However, as we have discussed, this approach faces challenges in processing sensory inputs of varying complexity and introducing unnecessary intricacies. 
Conversely, the Behavior Reflex model offers a simplified, yet effective, means of controlling autonomous vehicles by directly mapping sensory input to driving actions. 
While it provides real-time decision-making and computational efficiency, it grapples with challenges related to variations in human responses and limited environmental awareness.
Now, entering the fray is Direct Perception (DP), which navigates an intermediate path between Mediated Perception and Behavior Reflex. 
Direct Perception extracts critical affordance indicators from sensor data and maps them to driving actions. 
This approach offers adaptability to diverse driving conditions and scenarios, as it does not rely on an comprehensive World Model. 
However, DP also faces challenges in capturing all environmental nuances and handling complex traffic scenarios, particularly without a detailed understanding of the surroundings. 
As we progress in autonomous driving research and development, it is imperative to continue exploring these approaches' nuances, synergies, and limitations. 
Combining their strengths and mitigating their weaknesses will pave the way for safer, more efficient, and more intelligent autonomous vehicles, ultimately shaping the future of transportation.

%% file: sections/approach.tex
\section{Proposed Method}
\label{sec:proposed_method}
This section discusses the details of our proposed method, which consists of an ensemble neural network. 
We introduce the ``Mediated Perception Network," a core element of our architecture responsible for encoding and predicting embedding representations from camera images. 
We then explore the ``Behavior Reflex Network,'' which offers a starting point for our model and allows us to adapt and refine the policy as needed during autonomous driving. 
Next, we introduce the ``Direct Perception Network," a component inspired by human driving behavior. 
Finally, we present the ``Embedding Merging Approach" to merge the distinct embeddings generated by the networks mentioned above. 
We employ three merging methods involving convolution, direct concatenation, and self-attention to form a feature vector for the \gls{apn}. 

\subsection{Mediated Perception Network}
\begin{figure}[t!]
	\centering
	\includegraphics[width=\linewidth]{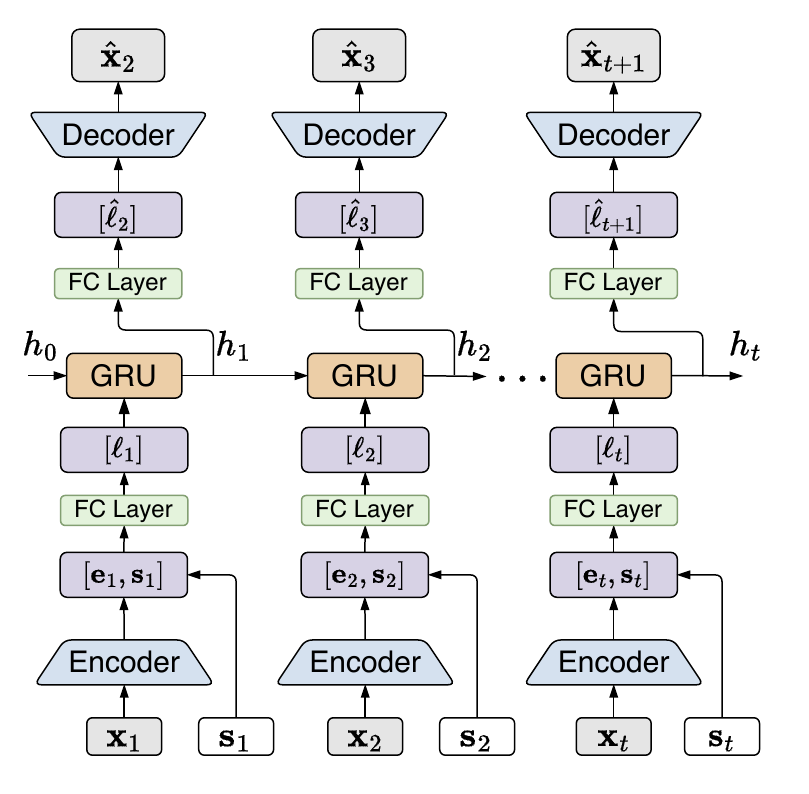}
	\caption{Mediated Perception neural-network architecture. A sequence of images is encoded into embedding space, fused (concatenated) with corresponding sensor measurements, then propagated through the recurrent network and outputs the estimate of both embedding states at timestep $t$.}
	\label{fig:merged_architecture_embedding}
\end{figure}
The conceptual framework of the Mediated Perception model typically comprises an Autoencoder (AE) designed to acquire embedding representations of the surrounding environment from a multitude of sensory inputs. This embedding representation is then forwarded through a Recurrent Neural Network (RNN) to capture and model temporal dependencies.
However, contemporary research endeavors have begun exploring neural network architectures incorporating fundamental physics principles within end-to-end deep learning models. In this context, we have adopted a physics-infused neural network architecture: the \gls{KARNet}, as introduced in Manjunatha et al.~\cite{manjunatha2023karnet}.
The \gls{KARNet} architecture is instrumental in acquiring two distinct embedding vector embeddings, denoted as $\Bell_t$ and $\Bell_{t+1}$. Herein, $\Bell_t$ signifies the current state of the environment, while $\Bell_{t+1}$ signifies the anticipated future state. For conciseness in subsequent discussions, we shall refer to the \gls{KARNet} network as the \gls{mpn}.

The \gls{mpn} architecture is shown in Figure~\ref{fig:merged_architecture_embedding}. 
To show the flow of computation, let us consider a sequence of $n$ consecutive frames taken from the front camera of a moving vehicle. 
Given the first five frames $\mathbf{x}_0,\dots\mathbf{x}_n$, their embedding representations $\Bell_0,\dots\Bell_n$ and corresponding sensor data $\mathbf{s}_0,\dots\mathbf{s}_n$, we aim to predict $\mathbf{x}_{n+1}$, $\Bell_{n+1}$ and $\mathbf{s}_{n+1}$ accordingly. As shown in Figure~\ref{fig:merged_architecture_embedding}, the image $\mathbf{x}_i$ is encoded into embedding space $\Bell_i=E(\mathbf{x}_i)$, here $E$ is the encoder. 
Note that the encoder/decoder can be any architecture or a typical convolutional neural network; hence, we assume a general architecture. The encoded vector $\Bell_i$ is used as an input for the RNN block whose output is next time step embedding space vector $\Bell_{i+1|i}$. The RNN block uses the general GRU formulation~\cite{chung2014empirical} and the predicted $\Bell_{i+1|i}$ is used as a hidden state for the next time step as well as an input for reconstructing the next time step image $\mathbf{x}_{i+1} = D(\Bell_{i+1|i})$ where $D$ is a decoder. Thus, the predicted embedding vector $\Bell_{i+1}$ at time step $i+1$ not only depends on $\Bell_i$, but also on $\Bell_{i-1}$. 

The overall loss function for a single-step prediction is given by equation!\ref{eq:ms-ssim}.

\begin{equation}
	L_{\text{\tiny{MS-SSIM}}} = \left[l_M(\mathbf{x}, \mathbf{y})\right]^{\alpha_M} \prod_{j=1}^{M} \left[c_j(\mathbf{x}, \mathbf{y})\right]^{\beta_j} \left[\mathbf{s}_j(\mathbf{x}, \mathbf{y})\right]^{\gamma_j},
	\label{eq:ms-ssim}
\end{equation}

\noindent where $\mathbf{x}$ and $\mathbf{y}$ are the images being compared,  $c_j(\mathbf{x}, \mathbf{y})$ and $\mathbf{s}_j(\mathbf{x}, \mathbf{y})$ are the contrast and structure comparisons at scale $j$, and the luminance comparison $l_M(\mathbf{x}, \mathbf{y})$ is computed at a single scale $M$, and $\alpha_M$ and $\beta_j$, $\gamma_j$ ($j=1,\ldots,N$) are weight parameters that are used to adjust the relative importance of the aforementioned components, i.e., contrast, luminance, and structure. These parameters are left to their default implementation values. 
More detailed information on the \gls{KARNet} architecture and hyper-parameters can be found in~\cite{manjunatha2023karnet}.

\begin{table}[h!]
    \renewcommand{\arraystretch}{1.25}
    \centering
    \caption{Autoencoder structure of \gls{KARNet}}
    \label{tab:autoencoder_architecture}
        \begin{tabular}{|c|c|c|c|}
        
        \multicolumn{2}{c}{\textbf{Encoder}}                                                                   & \multicolumn{2}{c}{\textbf{Decoder}}            \\ \hline
        \textbf{Layer}                                                       & \textbf{Output Shape} & \textbf{Layer}                                                         & \textbf{Output Shape}    \\ \hline
        \textbf{Input}                                                       & 1$\times$256$\times$256    & \textbf{Input}                                                         & 1$\times$128     \\ \hline
        \begin{tabular}[c]{@{}l@{}}conv3-2\\ conv3-2\end{tabular}   & 2$\times$128$\times$128    & \begin{tabular}[c]{@{}l@{}}tconv3-64\\ tconv3-64\end{tabular} & 64$\times$4$\times$4    \\ \hline
        \begin{tabular}[c]{@{}l@{}}conv3-4\\ conv3-4\end{tabular}   & 4$\times$64$\times$64      & \begin{tabular}[c]{@{}l@{}}tconv3-32\\ tocnv3-32\end{tabular} & 32$\times$8$\times$8    \\ \hline
        \begin{tabular}[c]{@{}l@{}}conv3-8\\ conv3-8\end{tabular}   & 8$\times$32$\times$32      & \begin{tabular}[c]{@{}l@{}}tconv3-16\\ tconv3-16\end{tabular} & 16$\times$16$\times$16  \\ \hline
        \begin{tabular}[c]{@{}l@{}}conv3-16\\ conv3-16\end{tabular} & 16$\times$16$\times$16     & \begin{tabular}[c]{@{}l@{}}tconv3-8\\ tconv3-8\end{tabular}   & 8$\times$32$\times$32   \\ \hline
        \begin{tabular}[c]{@{}l@{}}conv3-32\\ conv3-32\end{tabular} & 32$\times$8$\times$8       & \begin{tabular}[c]{@{}l@{}}tconv3-4\\ tconv3-4\end{tabular}   & 4$\times$64$\times$64   \\ \hline
        \begin{tabular}[c]{@{}l@{}}conv3-64\\ conv3-64\end{tabular} & 64$\times$4$\times$4       & \begin{tabular}[c]{@{}l@{}}tconv3-2\\ tconv3-2\end{tabular}   & 2$\times$128$\times$128 \\ \hline
        conv3-128                                                   & 128$\times$1$\times$1      & tconv3-1                                                      & 1$\times$256$\times$256 \\ \hline
        \end{tabular}
\end{table}

\subsection{Behavior Reflex Network}
\begin{figure}[t!] 
	\centering
	\includegraphics[width=\linewidth]{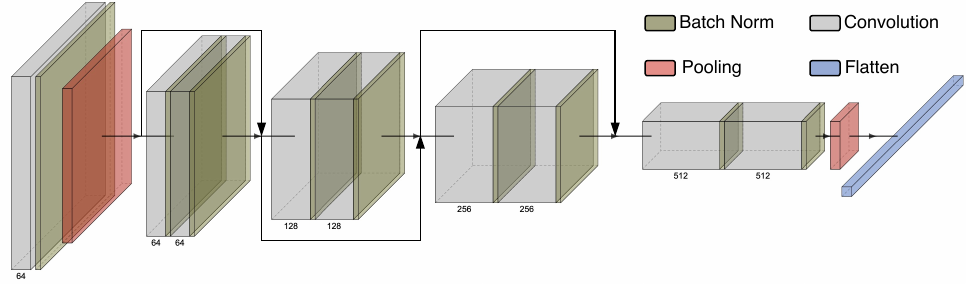}
	\caption{Behavior reflex network consists of four residual blocks with skip connection. The final flattened embedding is used as an input to the \gls{apn}.}
	\label{fig:behavior_reflex_network}
\end{figure}

\renewcommand{\arraystretch}{1.25}
\begin{table}
    \centering
    \caption{Behavior Reflex Model Structure}
    \label{tab:behavior_reflex_parameters}
    \begin{tabular}{|c|c|c|} 
    \hline
    \textbf{Layer Name }& \textbf{Output Shape} & \textbf{Parameters} \\ \hline 
    conv1 & $64 \times 128 \times 128$ & $7 \times 7,64$, stride 2 \\
    \hline 
    pool1 & $64\times64\times64$ & $3 \times 3$ max pool, stride 2 \\
    \hline 
    block1 & $256\times64\times64$ & {$\left[\begin{array}{l}3 \times 3,64 \\ 
    3\times3,\ 64\end{array}\right] \times 1$} \\
    \hline
    block2 & $512\times32\times32$ & {$\left[\begin{array}{l}3 \times 3,128 \\
    3\times3,\ 128\end{array}\right] \times 1$} \\
    \hline 
    block3 & $1024\times16\times 16$ & {$\left[\begin{array}{l}3 \times 3,\ 256 \\
    3\times3,\ 256\end{array}\right] \times 1$} \\ 
    \hline 
    block4 & $2048\times8\times8$ & {$\left[\begin{array}{l}3 \times 3,\ 512 \\
    3\times3,\ 512\end{array}\right] \times 1$} \\
    \hline 
    pool2 & $2048\times 1 \times1$ & $1 \times 1$ adaptive pool\\
    \hline
    fully connected & $128 \times 1$ & $2048\times 128$\\
    \hline
    \end{tabular}
\end{table}

For the \gls{brn}, we have utilized a residual neural network (ResNet) architecture (Figure~\ref{fig:behavior_reflex_network}). 
The \gls{brn} architecture first consists of a convolution operation and is followed by four residual block operations with skip connection between the blocks (Table~\ref{tab:behavior_reflex_parameters}).
Each residual block consists of two convolution operations with batch normalization after the convolution operation. 
After the residual block operation, we employed adaptive average pooling and a fully connected layer to produce the embedding $\eta_t$ of size $128 \times 1$. 
The \gls{brn} serves two purposes: a) a good initial policy (compared to random initialization) for warm starting learning, and b) it facilitates easy re-training for newer tasks, akin to the concept of residual policy learning~\cite{silver2018residual}. 
Here, we start with a fixed policy and then learn a residual policy to modify the fixed policy for a more complex situation. 

\subsection{Direct Perception Network}
\begin{figure}[t!] 
	\centering
	\includegraphics[width=\linewidth]{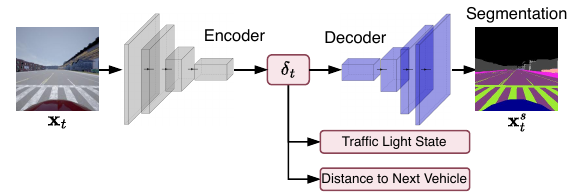}
	\caption{Direct Perception network consists of an encoder and a decoder where the output is a segmented image of the input image $\mathbf{x}_t$. Moreover, the embedding vector $\delta_t$ is also used to predict the distance to the next vehicle and traffic light condition.}
	\label{fig:direct_perception_network}
\end{figure}

In autonomous driving, end-to-end learning often proves inadequate for generating effective driving policies, neglecting the intricate task breakdown observed in human drivers. 
Rather than directly handling throttle and steering, humans instinctively employ a more complex hierarchical approach. 
They prioritize staying within lanes, maintaining a safe distance from the vehicle ahead, and gradually slowing down in response to a red traffic light.
The Direct Perception approach mirrors this human-inspired pipeline for the purpose autonomous driving policy development. 
To mirror the hierarchical approach used by humans, the Direct Perception models predicts the auxiliary tasks such as distance to the front vehicle, distance to curbs or centre lane, traffic light condition, speed of the front vehicle, etc. and use this information to predict the driving actions.
These auxiliary tasks yield a remarkably feature-rich embedding vector.

In our study, as depicted in Figure~\ref{fig:direct_perception_network}, the \gls{dpn} is trained to perform semantic segmentation of the driving environment while concurrently predicting auxiliary tasks. 
These auxiliary tasks involve estimating the distance to the front vehicle and traffic light state. 
Distance (measured in meters) to the front vehicle is discretized into five distinct classes: $[0-10]$, $[10-20]$, $[20-30]$, $[30-40]$, and $[40-100]$, thereby constituting a five-class classification problem. 
Traffic light status is cast as a binary classification approach with $0$ representing red or yellow lights and $1$ representing green light.
Lastly, the primary task of semantic segmentation entails a classification challenge encompassing $23$ distinct classes.
We have used the same encoder and decoder architecture as shown in Table~\ref{tab:autoencoder_architecture}. 
However, to facilitate semantic segmentation, the final layer of the decoder outputs $23$ channels instead of one channel. 
We have used the cross-entropy loss for each classification task to train the neural network.
By training in this manner, the same embedding vector $\delta_t$ is used to predict the semantic segmentation, distance to the front vehicle, and traffic light state. 

\subsection{Embedding Merging Approach}
\label{sec:embedding_merging_approach}
Given the distinct embeddings acquired from three neural networks: Mediated Perception ($\Bell_t, \Bell_{t+1}$), Behavior Reflex ($\eta_t$), and Direct Perception ($\delta_t$), we use three methods to merge the embedding to form a feature vector for \gls{apn}. 
The first method involves a convolution operation, while the second involves direct concatenation.
Finally, the third method uses a self-attention mechanism followed by averaging. 
We excluded alternative augmentation methods, such as averaging and element-wise multiplication, from our analysis. 
The rationale behind this exclusion is the potential loss of our capacity to discern and evaluate the individual significance of each modeling approach. 
By employing convolution, direct concatenation, and self-attention, we aim to preserve these core components' interpretability and discernible impact on the resulting embeddings.

\begin{figure}[th!]
	\centering
	\includegraphics[width=\columnwidth]{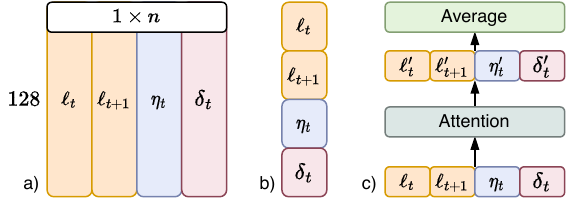}
	\caption{Merging methodologies: a) convolution operations across all the embedding embeddings. b) direct concatenation and c) self-attention mechanism followed by averaging.}
	\label{fig:merging}
\end{figure}

Figure~\ref{fig:merging}a) shows the convolution approach to merge the embeddings. 
The embeddings $\Bell_t, \Bell_{t+1}, \eta_t, \text{and}\ \delta_t$ are stacked horizontally, the results stack of tensor is of size $128 \times p$, where $p$ is the number of embeddings and the $128$ is the embeddings' length, a convolution operation with a kernel size $1 \times p$ is used which spans across the stacked tensor horizontally, so that results tensor $\alpha_t$ is of the size $128 \times 1$. 
The resultant $\alpha_t$ tensor is used as an input for the \gls{apn}. 
Note that for different combination of embeddings ($\Bell_t, \Bell_{t+1}, \eta_t, \text{and}\ \delta_t$), the convolution kernel size is adjusted appropriately. 
For example, when $\Bell_t, \Bell_{t+1}, \text{and}\ \eta_t$ are used (Mediated Perception and Behavior Reflex models), the kernel size is set to $1 \times 3$. Moreover, we also explored a convex-constrained convolution operation, i.e., where the sum of kernel weights is $1$, and an unconstrained method where the kernel weights can take any value. 

Figure~\ref{fig:merging}b) shows the stacking approach where the embeddings $\Bell_t, \Bell_{t+1}, \eta_t, \text{and}\ \delta_t$ are stacked vertically to form a tensor of size $128p \times 1$ where $p$ is the number of embeddings. 
Note that when the embeddings are stacked, scaling problems can occur because each embedding is learned from different modeling approaches. To compensate for the scaling issue, we use three fully connected layers with a gradual decrease in size before the action prediction.
To have the same size merged tensor as the convolution merging approach, the output size after three fully connected layers is $128$.
Finally, we also explored a self-attention mechanism between the embedding vectors as given by Equation~\ref{eq:self-attention}:

\begin{equation}       
    \label{eq:self-attention}
    I_{i j}=\sum_{a, b \in \mathcal{N}_{k}(i, j)} \operatorname{softmax}_{a b}\left(q_{i j}^{\top} k_{a b}\right) v_{a b},
\end{equation}

\noindent where, $q_{i j}=W_{Q} e_{i j}$, $k_{a b}=W_{K} e_{a b}$, and the values $v_{a b}=W_{V} e_{a b}$ are query, key, and values respectively, with learnable weights $W_{Q}, W_{K}, W_{V}$. 
Here $e_{i j}$ represents a pair of embedding vectors. After the self-attention mechanism, we employed an average operation to get an output ($\alpha_t$) of size $128 \times 1$, which is then used as an input for action prediction.

\subsection{Action Prediction Network}
\begin{figure}[t!] 
	\centering
	\includegraphics[width=\linewidth]{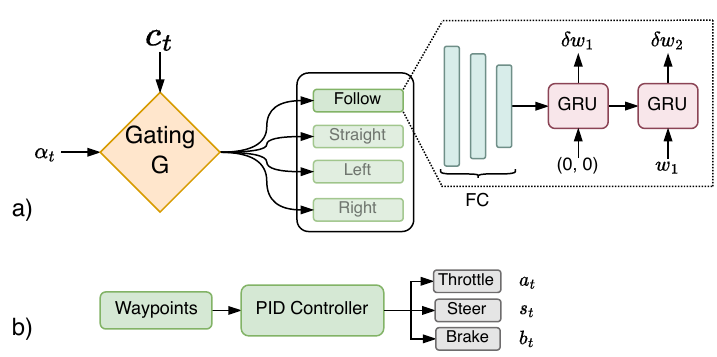}
	\caption{a) Action Prediction network architecture is used for predicting the waypoints. Each sub-network (Follow, Straight, Left, Right) consists of 3 layers of fully connected (FC) and a two-time step \gls{gru} network, which is used to predict the two waypoints by taking the vehicle's present location as the origin. b) A PID controller converts the waypoints to throttle/acceleration, steer, and brake.}
	\label{fig:action_network}
\end{figure}

For training the Action Prediction Network (\gls{apn}), we utilize the \gls{mpn}, \gls{brn}, and \gls{dpn} as features extracts where the features are the corresponding embeddings $\Bell_t, \Bell_{t+1}$, $\eta_t$, and $\delta_t$. 
It's important to note that the weights of \gls{mpn} and \gls{dpn} remain fixed, and only the weights of the \gls{apn} and \gls{brn} undergo training. 
This approach of training the \gls{brn} along with the \gls{apn} draws parallels with motor adaptation learning theory, which entails adapting the execution of a well-practiced action to sustain performance when faced with alterations in the environment or one's physical condition~\cite{krakauer2019motor}.
Moreover freezing the weights of other networks (\gls{mpn} and \gls{dpn})
and separately training the action prediction finds similarities with neuroscience, where pre-motor areas can be activated without resulting in motor function activation~\cite{hesslow2012current}. 

To facilitate goal-oriented behavior, we introduce the control command $c_{t}$, which guides the vehicle from an initial location to its final destination, similar to the approach presented by Liang et al.~\cite{liang2018cirl}. 
This command $c_{t}$ operates as a categorical variable, regulating selective branch activation through the gating function $G(c_{t})$. 
The available options for $c_{t}$ encompass four distinct commands: ``Follow the lane," ``Drive straight at the next intersection," ``Turn left at the next intersection," and ``Turn right at the next intersection.'' 
To enable action prediction, we specifically train four policy branches, each encoding unique hidden knowledge for its corresponding command. These policy branches are implemented as auto-regressive waypoint networks, utilizing GRUs~\cite{chitta2022transfuser} for their architecture (see Fig.~\ref{fig:action_network}).

To initiate the GRU network, we calculate the first hidden vector $h_0$ using the combined embedding vector $\alpha_t$, employing the function $h_0=\mathrm{FC}(\alpha_t)$, where $\mathrm{FC}$ represents a fully connected neural network with an output size of 64. The input for the GRU network consists of the previous waypoint's position. 
The GRU network's output corresponds to the difference between the current waypoint and the next waypoint, expressed as $w_i = w_{i-1} + \delta_t w_i$, with $w_{i-1}$ serving as the input to the GRU and $\delta_t w_i$ representing the prediction. 
Notably, the waypoints are forecasted with respect to the car's current position, starting from the reference point $w_0 = (0, 0)$. 
These waypoints are converted to steering, braking, and acceleration commands using a PID controller.
In this study, we consider two sets of waypoints $w_1$ at \qty{5}{\meter} from the ego vehicle and $w_2$ at \qty{10}{\meter}. 
Such a consideration is based on the empirical evidence from the two-point visual driver control model (TPVDCM)~\cite{salvucci2004two, okamoto2019vision}.
Inspired from these works and for simplicity, we will call $w_1$ as $w_{\text{near}}$ and $w_2$ as $w_{\text{far}}$ for the rest of the paper.

\section{Experiments}

\subsection{Simulated Data}

\begin{figure}[ht!]
    \centering
    \begin{subfigure}
        \centering
        \includegraphics[height=1.4in]{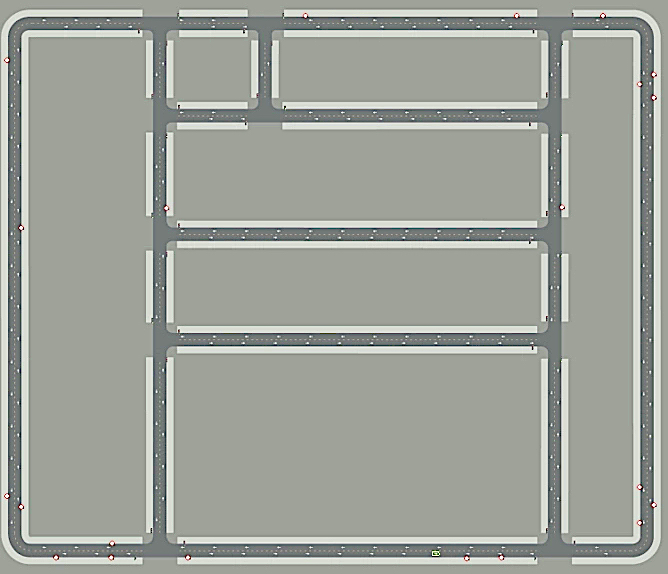}
    \end{subfigure}%
    ~ 
    \begin{subfigure}
        \centering
        \includegraphics[height=1.4in]{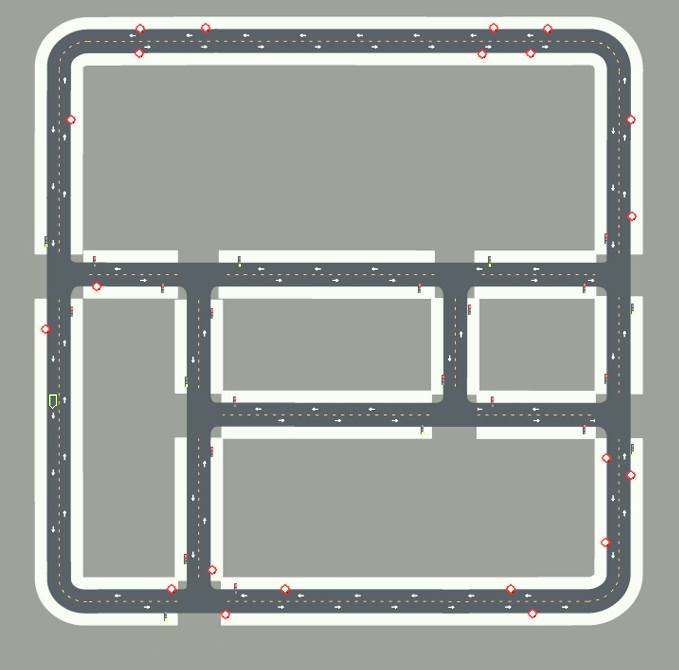}
    \end{subfigure}
    \caption{a) Map of the town in CARLA used for data generation. b) Map of the town used for testing.}
    \label{fig:town}
\end{figure}

The training data was generated using the CARLA simulator~\cite{dosovitskiy2017carla}. 
A total of 1.4M time steps were generated using random roll-outs (random starting and goal points on the map shown in Fig.~\ref{fig:town}a)) utilizing the internal CARLA vehicle autopilot. 
The simulated data includes four-directional camera images (front/left/right/rear) along with their corresponding semantic segmentation, IMU, waypoints for navigation and other sensor data (speed, steering, LIDAR, GNSS, etc.), desired control values, and additional experimental data for auxiliary tasks such as traffic light information and distance to the front vehicle.

\subsection{Metrics}  
\label{sec:metrics}

We utilize two key metrics to assess the agent's performance in a driving simulation: route completion (RC) and infractions per kilometer (IN/km). 
Our benchmark tests are conducted on the Town02 map (as shown in Figure~\ref{fig:town}b)) within the CARLA environment, exposing the agent to varying weather and daylight conditions not encountered during training. 
The agent's task involves navigating between two randomly selected points on the map, covering an average distance of \qty{0.65}{\kilo\meter} while sharing the road with $20$ vehicles and $20$ pedestrians. 
These metrics are determined as averages over fifteen runs. 
If the agent deviates off-road or collides with another vehicle or stationary object, we terminate the episode and calculate the relevant metrics.

\begin{itemize}   
	\item Route Completion (RC) quantifies the percentage of the route successfully covered between the designated starting and ending points. 
    Notably, the total route distance may differ across different experimental runs due to variations in the starting and ending locations. 
    Consequently, when presenting this metric, we calculate the average value to account for such disparities.
	\item Infractions per kilometer (IN/km) measures the number of infractions (including collisions with vehicles, pedestrians, lane deviations, and off-road excursions) in relation to the total kilometers traveled. This is computed as:
	      \begin{equation}
	      	\begin{aligned}
	      		\text{Infraction per km} = \frac{\sum \# \text{Infractions}}{K} 
	      	\end{aligned}
	      \end{equation}
	      Where $K$ represents the total distance covered in the specified route.    
\end{itemize}

%% file: sections/results.tex
\section{Results}
In this section, we delve into evaluating the ensemble neural network (i.e., \gls{mpn}+\gls{brn}+\gls{dpn}). 
Our objective is to assess the network's performance and shed light on each constituent model's distinct contributions under varying driving conditions. 
Additionally, we explore the impact of different merging approaches on these ensemble architectures.
Our analysis employed three merging approaches: convolution, direct concatenation, and self-attention. 
Among these, our results demonstrate that the convolution operation outperforms the other two approaches, as evidenced by its superior route completion rates and fewer infractions per kilometer.
Hence, in the subsequent sections, we will discuss the convolution merging method and its variations. 
For a more detailed examination of our merging approach, please refer to Section~\ref{sec:embedding_merging_approach}, where we provide insights into unconstrained and constrained convolution approaches.

\subsection{General Performance Analysis}
Table~\ref{tab:combined_results_unconstrained} showcases the performance of various models in a route completion task, measured as the percentage of completed routes and the number of infractions per kilometer with an unconstrained merging approach. 
To serve as baseline we trained only the \gls{brn} which resulted in 74.56\% route completion and 41.32 IN/km.
The \gls{dpn}+\gls{brn}+\gls{mpn} model outperformed them all, achieving an impressive 98.85\% route completion rate and 26.49 infractions per kilometer, followed by the \gls{dpn}+\gls{brn} and \gls{mpn}+\gls{brn} model. 

\begin{table}[htbp]
	\renewcommand{\arraystretch}{1.5}
	\centering
	\caption{Bench-marking results using Unconstrained Merging.}
	\begin{tabular}{p{20mm}cc}
		\hline
		\textbf{Model}        & \textbf{Route Completion (\%)} & \textbf{No. of Infractions/Km}\\
		\hline
		\textbf{\gls{mpn}+\gls{brn}}      & 88.34 & 34.53\\
		\textbf{\gls{dpn}+\gls{brn}}      & 96.17 & \textbf{25.17}\\
		\textbf{\gls{dpn}+\gls{brn}+\gls{mpn}} & \textbf{98.85} & 26.49\\
		\hline
	\end{tabular}
	\label{tab:combined_results_unconstrained}
\end{table}

It's noteworthy that \gls{dpn}'s contribution to reducing infractions is evident, with the \gls{dpn}+\gls{brn} model displaying the least infractions per kilometer at 25.17. 
This can be attributed to the \gls{dpn}'s specialized training for predicting critical road conditions, such as traffic light conditions and distance to front vehicles, which directly impact driving safety.
Moreover, the positive influence of \gls{dpn} extends to the \gls{dpn}+\gls{brn}+\gls{mpn} model also. 
This suggests that \gls{dpn}'s capabilities significantly contribute to the combined model's safety.
In contrast, the \gls{mpn}+\gls{brn} combination exhibited higher infractions and a lower route completion rate. 
The absence of safety-related information within the \gls{mpn}+\gls{brn} model, which sets it apart from the \gls{dpn}, can be identified as the primary reason for this outcome. 
The infractions in this context include not only minor violations, such as lane violations but also more severe offenses like veering off the road. As a result, both the route completion rate and infractions remain notably high.
It is important to note that when the \gls{mpn} is integrated alongside both \gls{brn} and \gls{dpn}, the result is a notable improvement in performance. This integration delivers the most favorable outcomes across the evaluated models.

\begin{table}[htbp]
	\renewcommand{\arraystretch}{1.5}
	\centering
	\caption{Bench-marking results using Constrained Merging.}
	\begin{tabular}{p{20mm}cc}
		\hline
		\textbf{Model}        & \textbf{Route Completion (\%)} & \textbf{No. of Infractions/Km}\\
		\hline
		\textbf{\gls{mpn}+\gls{brn}}      & 85.98 &  40.89\\
		\textbf{\gls{dpn}+\gls{brn}}      & 92.79 & \textbf{29.81}\\
		\textbf{\gls{dpn}+\gls{brn}+\gls{mpn}} & \textbf{97.75} &  35.65\\
		\hline
	\end{tabular}
	\label{tab:combined_results_constrained}
\end{table}
 
Table~\ref{tab:combined_results_unconstrained} showcases the performance of various models with a constrained merging approach. 
Comparing the two scenarios, unconstrained merging generally outperforms constrained merging in terms of route completion and infraction per kilometer, with the \gls{dpn}+\gls{brn}+\gls{mpn} combination delivering the best overall results in both cases. 
One plausible hypothesis to explain the enhanced performance in unconstrained merging is that the weights are not constrained, allowing them to adapt based on the scale of the embeddings. 
This adaptability likely leads to better model performance. 
Future research could delve deeper into the mechanisms behind this adaptability and its impact on the merging process.

A plausible explanation for the enhanced performance of the \gls{dpn}+\gls{brn}+\gls{mpn} model can be attributed to the over-parameterization of neural networks~\cite{liu2022loss}. 
Empirical observations suggest an intriguing paradox: deep learning models often exhibit good generalization and lower error rates in test cases despite having significantly more parameters than training examples \cite{zhang2021understanding}. 
This empirical observation appears to contradict traditional learning theories \cite{arora2018stronger}. 
As a result, substantial effort has been invested in investigating the theoretical properties of deep learning models \cite{kawaguchi2022generalization, neyshabur2017exploring, arora2018stronger, wu2017towards, nagarajan2021explaining}. 
Recently, Casper et al.,  \cite{casper2021frivolous} explored the emergence of prunable and redundant units in relation to the generalization ability of deep neural networks. 
Their observations indicate that prunable and redundant units proliferate at a rate exceeding the model size. Based on these findings, they put forth the following hypothesis: consider a narrow, deep network $\mathcal{N}$ and a wide one $\mathcal{W}$, both sharing the same architecture, training data, initialization, hyperparameters, and training procedure. 
In such a scenario, $\mathcal{W}$ develops a higher proportion of prunable and/or redundant units compared to $\mathcal{N}$ while maintaining equal or better generalizability.
In our case, as we introduce more sub-networks into the architecture, the number of parameters increases, thus leading to overparameterization and, in turn, better performance. 
Moreover, the information captured by different embeddings ($\Bell_t, \Bell_{t+1}, \eta_t, \text{and}\ \delta_t$) can overlap, leading to redundancy, in line with the aforementioned hypothesis. 
However, a notable benefit of our approach lies in the principles nature of this overparameterization, arising from the use of three distinct networks rather than a single neural network. 
This structured approach permits a detailed examination of each network's contribution, which we delve into in the subsequent section. 
Such scrutiny may not be feasible with a single neural network architecture.

\subsection{Ablation Analysis (Embeddings)}
Ablation analysis is a valuable method for deconstructing complex systems to gain insights into their inner workings. 
The technique involves systematically disabling or removing specific components, features, or variables from a neural network to assess their individual contributions and their impact on overall performance.
To evaluate the importance of each modeling approach, namely Mediated Perception, Behavior Reflex, and Direct Perception, in the ensemble network, we can utilize the ``Feature Layer Ablation" functionality available in the Captum package~\cite{kokhlikyan2020captum}.
The importance of each embedding, and consequently, each modeling approach, can be determined by ablating the embedding and computing the absolute error in predicting the waypoints, as defined by Equation~\ref{eq:ablation} below.

\begin{equation}
    |P(e) - P(e_i)| = |\Delta x|_s + |\Delta y|_s.
    \label{eq:ablation}
\end{equation}

In this equation, $P(e)$ represents the predicted waypoints with all embeddings $e$, and $P(e_i)$ is the prediction with the embedding $e_i$ set to a zero tensor. As an example, in the ensemble model with embeddings $\Bell_t, \Bell_{t+1}, \eta_t, \text{and}\ \delta_t$, we substitute the $\Bell_t$ tensor with zeros and compute the absolute error in the waypoints predictions.

It is important to note that we predict two sets of waypoints, $w_{\text{near}}$ and $w_{\text{far}}$, with reference to the ego vehicle's position $s$. This results in two ablation values corresponding to the two waypoints. 
Moreover, we calculate the ablation value, as per Equation~\ref{eq:ablation}, along three different paths: straight, right turn, and left turn. The average contribution can then be calculated using

\begin{equation}
    T(e_i, s_i, s_f) = \frac{1}{m}\sum_{s=s_i}^{s_f} |\Delta x|_s + |\Delta y|_s.
    \label{eq:avg_ablation}
\end{equation}

Here, $s_i$ and $s_f$ represent the initial and final positions, and $m$ is the total number of path points between those positions.

\begin{figure*}[ht!]
    \centering
    \begin{subfigure}[]
        \centering
        \includegraphics[height=1.70in]{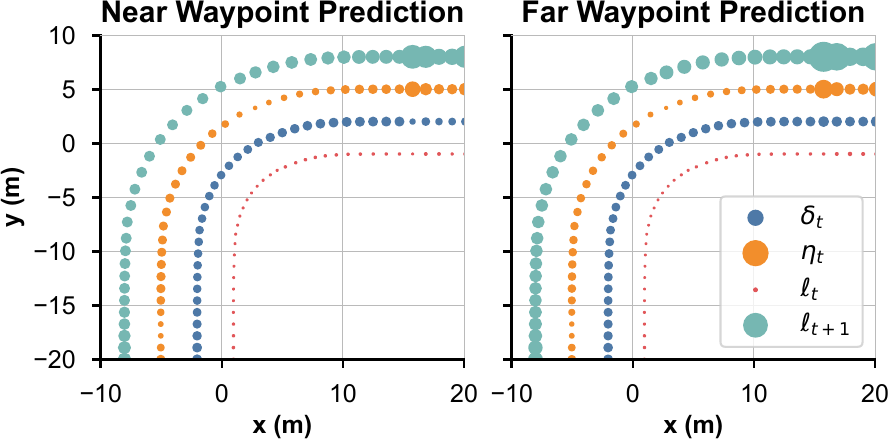}
    \end{subfigure}
    \begin{subfigure}[]
        \centering
        \includegraphics[height=1.70in]{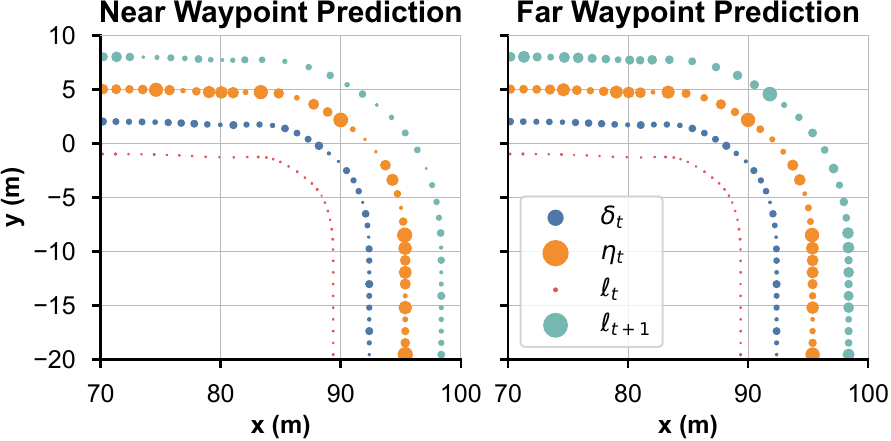}
    \end{subfigure}
    \caption{Ablation values for each embedding $\delta_t$, $\eta_t$, $\Bell_t$, and $\Bell_{t+1}$ plotted along the path a) turning right and b) turning left.}
    \label{fig:ablation-analysis-plot}
\end{figure*}

\begin{table}[htbp]
\renewcommand{\arraystretch}{1.5}
	\centering
	\caption{Ablation results using Unconstrained Merging.}
    \begin{tabular}{llrrrr}
    \hline
    \multirow{2}[4]{*}{\textbf{Path Type}} & \multicolumn{1}{l}{\multirow{2}[4]{*}{\textbf{Prediction}}} & \multicolumn{4}{c}{\textbf{Embedding}} \\
\cline{3-6}      &   & $\delta_t$ & $\eta_t$ & $\Bell_t$ & $\Bell_{t+1}$ \\
    \hline
    \multirow{2}[2]{*}{Left Turn} & Short term & 0.1164 & 0.370 & 0.0034 & 0.237 \\
      & Long term & 0.0896 & 0.382 & 0.0037 & 0.306\\
    \hline
    \multirow{2}[2]{*}{Right Turn} & Short term & 0.129 & 0.310 & 0.0039 & 0.237 \\
      & Long term  & 0.093 & 0.330 & 0.0035 & 0.364 \\
    \hline
    \multirow{2}[2]{*}{Straight} & Short term & 0.124 & 0.384 & 0.0049 & 0.234 \\
      & Long term  & 0.118 & 0.358 & 0.0046 & 0.407\\
    \hline
    \end{tabular}%
  \label{tab:ablation-analysis}%
\end{table}%

Since the unconstrained combined model (\gls{dpn}+\gls{brn}+\gls{mpn}) resulted in the best performance, we will only discuss the ablation analysis in this context. 
Table~\ref{tab:ablation-analysis} provides the importance values of four different embeddings ($\delta_t$, $\eta_t$, $\Bell_t$, and $\Bell_{t+1}$) for predicting waypoints in different path types (left, right and straight) and time frames (Short term and Long term). 
$\eta_t$ (Behavior Reflex Network) is consistently the most critical embedding in both short-term and long-term predictions for all path types in determining the chosen path. 
$\Bell_{t+1}$ (Mediated Perception Network) is consistently the second most crucial embedding for all path types in short-term and long-term predictions. This suggests that anticipating the traffic state is vital for predicting the waypoints.
Moreover, the differences in ablation values between the embeddings for different path types are relatively small (except for $\Bell_{t+1}$), suggesting that the same embeddings are relevant for predicting waypoints across different turning scenarios.
Interestingly, the difference in ablation values for short-term and long-term prediction is highest in $\Bell_{t+1}$, which signifies the importance of $\Bell_{t+1}$ in long-term prediction.
While $\delta_t$ and $\Bell_t$ are less influential, they still contribute to the prediction, albeit to a minor degree when used with other embeddings.
Nonetheless, the bench-marking results (Table~\ref{tab:combined_results_unconstrained} and \ref{tab:combined_results_constrained}) indicate that $\delta_t$ directly influences the number of infraction, thus contributing to the safety. 
Figure~\ref{fig:ablation-analysis-plot} shows the ablation values for each embedding along the path, taking a left and right turns.

Concretely, the \gls{brn} serves as the foundation network, setting the baseline performance standards. 
Nevertheless, it lacks the essential attributes of safety assurance and future prediction capabilities. 
This is where the \gls{dpn} comes into play, offering crucial safety information while the \gls{mpn} contributes insights into future occurrences.
When combined, the \gls{brn} and \gls{dpn} alone already exhibit superior performance, marked by increased route completion rates and decreased infraction rates. 
However, adding the \gls{mpn} to the \gls{brn}+\gls{dpn} configuration can further enhance route completion.

%% file: sections/conclusion.tex
\section{Conclusion}
We conducted an extensive investigation to assess the effectiveness of ensemble modeling strategies in autonomous driving. 
Our work introduced a neural network architecture that combines multiple deep learning approaches for autonomous driving, specifically the Behavior Reflex Network (\gls{brn}), Mediated Perception Network (\gls{mpn}), and Direct Perception Network (\gls{dpn}). 
We focused on evaluating these approaches' utility rather than introducing novel neural network designs.

The \gls{brn}, trained solely using the front-camera images through imitation learning, was designed to capture the immediate environmental state and compress it into an embedding denoted as $\eta_t$. 
The \gls{mpn} was responsible for predicting both the present and future states of traffic using latent representations $\Bell_{t}$ and $\Bell_{t+1}$, drawing inspiration from the concept of ``World Models." 
Meanwhile, the \gls{dpn} learned to predict auxiliary tasks, including segmenting input front-camera images, estimating the distance to the front vehicle, and assessing traffic light conditions using a common embedding, $\delta_t$. 
These four embeddings, $\Bell_t$, $\Bell_{t+1}$, $\eta_t$, and $\delta_t$, formed the foundational models representing various aspects of autonomous driving and were combined to create a feature vector for action prediction.

To merge these embeddings effectively, while preserving interpretability and distinguishing their impact on the final results, we employed three different methods: convolution weighting, direct stacking, and self-attention. 
Among the three merging methods, convolution weighting provided the best performance.
Our tests were conducted in CARLA, featuring various maps, weather conditions, and daylight settings not encountered during training. The agent's task was to navigate between two randomly selected points on the map, covering an average distance of 0.65 kilometers while sharing the road with 20 vehicles and 20 pedestrians.

In summary, the \gls{brn} served as the foundation network, establishing baseline performance standards but lacked safety assurance and future prediction capabilities. While the \gls{dpn} bolstered safety by providing crucial information, the \gls{mpn} offered insights into future traffic states. Combining \gls{brn} and \gls{dpn} notably improved performance, and adding the \gls{mpn} further enhanced it, showcasing the effectiveness of ensemble modeling for diverse driving conditions

This research enhances our understanding of how distinct modeling strategies can be effectively utilized in autonomous driving. 
While our study primarily focused on supervised learning, the integration of reinforcement learning (RL) presents an intriguing direction for future research. 
By incorporating RL techniques, we can explore how the ensemble modeling strategies we've examined in this research interact with adaptive learning algorithms. 
This could lead to even more robust and adaptable autonomous driving systems. 
Further, we plan to move beyond controlled simulation environments like CARLA to the deployment of autonomous driving systems in real-world scenarios. This involves addressing issues related to safety and the complexities of diverse, uncontrolled traffic conditions.